\documentclass[conference]{IEEEtran}
\IEEEoverridecommandlockouts
\usepackage{cite}
\usepackage{amsmath,amssymb,amsfonts}
\usepackage{graphicx}
\usepackage{textcomp}
\usepackage{xcolor}

\usepackage{booktabs}
\usepackage{mathtools}
\usepackage{multirow}
\usepackage{pgfplots}
\pgfplotsset{compat=1.15}
\usepackage{pgfplotstable}
\usepackage{ragged2e}
\usepackage{subcaption}

\usepackage[algo2e]{algorithm2e}
\usepackage{algpseudocode}
\usepackage{algorithm}

\def\BibTeX{{\rm B\kern-.05em{\sc i\kern-.025em b}\kern-.08em
    T\kern-.1667em\lower.7ex\hbox{E}\kern-.125emX}}
\begin{document}

\title{LSTM-Autoencoder based Anomaly Detection for Indoor Air Quality Time Series Data\\

}

\author{\IEEEauthorblockN{Yuanyuan Wei\textsuperscript{1}, Julian Jang-Jaccard\textsuperscript{1}, Wen Xu\textsuperscript{1}, Fariza Sabrina\textsuperscript{2}, Seyit Camtepe\textsuperscript{3}, Mikael Boulic\textsuperscript{4}}
	\IEEEauthorblockA{\textit{} \textit{}
		\textsuperscript{1}CybersecurityLab, Comp Sci/Info Tech, Massey University, Auckland, 0632, NEW ZEALAND \\
		\textsuperscript{2}School of Engineering and Technology, Central Queensland University, Sydney NSW 2000, AUSTRALIA \\
		\textsuperscript{3}CSIRO Data61, AUSTRALIA \\
		\textsuperscript{4}School of Built Environment, Massey University, Auckland, 0632, NEW ZEALAND\\
		y.wei1@massey.ac.nz, j.jang-jaccard@massey.ac.nz, w.xu2@massey.ac.nz, f.sabrina@cqu.edu.au,\\ Seyit.Camtepe@data61.csiro.au, m.boulic@massey.ac.nz}}

\maketitle

\begin{abstract}
Anomaly detection for indoor air quality (IAQ) data has become an important area of research as the quality of air is closely related to human health and well-being. However, traditional statistics and shallow machine learning-based approaches in anomaly detection in the IAQ area could not detect anomalies involving the observation of correlations across several data points (i.e., often referred to as long-term dependences). We propose a hybrid deep learning model that combines LSTM with Autoencoder for anomaly detection tasks in IAQ to address this issue. In our approach, the LSTM network is comprised of multiple LSTM cells that work with each other to learn the long-term dependences of the data in a time-series sequence. Autoencoder identifies the optimal threshold based on the reconstruction loss rates evaluated on every data across all time-series sequences. Our experimental results, based on the Dunedin $CO_2$ time-series dataset obtained through a real-world deployment of the schools in New Zealand, demonstrate a very high and robust accuracy rate (99.50\%) that outperforms other similar models.
\end{abstract}


\section{Introduction}
\label{sec:introduction}
\IEEEPARstart{T}{he} indoor air quality (IAQ) is closely related to human health, productivity, and work efficiency\cite{kim2014issaq}. Good air quality is even more important for children who spend a vast majority of their time at school. Providing children with fresh air in their classroom environment is of high importance for their health and well-being. However, the formulation of $CO_2$, which is considered to be the major constituents of indoor air pollutants, can be easily built up by children studying/playing inside classrooms and also accompanied by the emission from floors and other surface~\cite{motlagh2019indoor}. Such $CO_2$ can become the basis of creating harmful mold and bacteria that could contribute to poor health and degrading academic performance.

Constant monitoring of indoor air quality including the measurement of the level of $CO_2$ in school environments has been problematic for many countries including the OECD as the large-scale monitoring of indoor air quality has proven to be too expensive for budget-strapped schools. To address this concern, a team of researchers at Massey University developed a low-cost monitoring suite called SKOol MOnitoring BOx (SKOMOBO) with the National Institute of Water and Atmospheric Research (NIWA) \cite{wang2017integrating, weyers2017low, wang2018deployment}.  A SKOMOBO unit is a small box, approximately the size of 100 × 100 × 100 mm, designed to house a number of low-cost sensors that could capture a number of indoor air quality-related data such as particulate matter ($PM_{2.5}$ and $PM_{10}$), temperature, relative humidity, carbon dioxide ($CO_2$) and human occupancy in classrooms.

The monitoring of measurable indoor air quality is challenging due to the fluctuation of IAQ data reading and the question of data quality, for example, deploy in non-stationary or uncontrolled environments ~\cite{zeng2022air}. In addition, rare or consistent contamination events can also corrupt the data quality (i.e., natural disasters such as fire, flooding, thunderstorms). As a result, numerous proposals for detecting anomalous events in IAQ have been attempted by utilizing the latest advancement in artificial intelligence-based techniques. For instance, many attempted to use statistical methods (e.g., using the means, standard deviation, Gaussian q-distribution) \cite{wei2020large, wu2019lstm} and other combinations of shallow machine learning techniques (e.g., kNN, k-means, regression) \cite{ottosen2019outlier, li2021clustering, zhou2020variational} to find the patterns of normal behaviors of IAQ and use that as a basis to detect anomalous events or for improving forecasting capabilities such as \cite{sharma2021indoairsense, rastogi2019internet}.  However, these existing methods could not detect anomalies where the observation of several correlated data points is necessary. 

In this research, we propose a hybrid deep learning model that combines the capabilities of long short-term memory (LSTM) and Autoencoder (AE) for detecting anomalous data points in IAQ datasets based on the understanding of long-term dependencies that exist in data samples.

The main contributions of our proposed model are the following. 

\textbf{Summary of Original Contributions}
 
\begin{itemize}
  	\item In our proposed model, LSTM networks are comprised of multiple LSTM units that work with each other to learn the long-term correlation of data within a time series sequence. Autoencoder is used to identify the optimal threshold based on the reconstruction error rates evaluated on every data across all time-series sequences. This threshold is used to identify anomalies.
  	\item We apply our proposed model to the Dunedin $CO_2$ Dataset obtained from a real-world deployment of multiple primary/secondary schools in New Zealand.  
  	\item We compare the performance of the proposed model with other similar approaches that use different aspects of LSTM and/or AE. Our experimental results, performed based on the comprehensive set of evaluation criteria, demonstrate that our proposed model can effectively detect anomalies reaching the detection accuracy that exceeds 99\%.
\end{itemize}
 
The rest of this paper is structured as follows. Section~\ref{sec:rw} introduces related works in the field of indoor air quality.  Section~\ref{sec:method} introduces the details of our proposed model. Section~\ref{sec:re} illustrates the experimental setup and analysis of results evaluated on the Dunedin $CO_2$ dataset. Section~\ref{sec:Conclusion} concludes the paper with the planned future works.

\section{Related Work}\label{sec:rw}
We review the existing state-of-the-art techniques for detecting anomalies in indoor air quality datasets and other similar fields.

Ottosen et al.~\cite{ottosen2019outlier} introduced two anomaly detection techniques that use k-Nearest Neighbour (KNN) and AutoRegressive Integrated Moving Average (ARIMA) to detect both point and contextual anomalies separately in a low-cost air quality dataset. In their approach to detect point anomalies, they simply utilized the average Euclidean distance to compute the similarity with the remaining points and assign an anomaly score to the individual point. ARIMA was used to detect contextual anomalies by calculating anomaly scores for each data point between the model and measurement based on the absolute value of the residual. Both point and contextual anomalies are classified into two clusters as normal and anomaly by K-means clustering. \\
Wei et al.~\cite{wei2020large} proposed a hybrid model of MSD-Kmeans to detect anomalies with indoor PM$_{10}$ dataset. They first used the statistical method of Mean and Standard Deviation (MSD) which was used to eliminate noisy data to reduce the impact of clustering from the noise. Then they applied the K-means algorithm to achieve better local optimal clustering. The performance of their proposal showed the detection accuracy (97.6\%) and F1-score (91.9\%).\\
Li et al.~\cite{li2021clustering} proposed clustering-based Fuzzy C-means. In their approach, the authors used a reconstruction criterion to reconstruct the optimal cluster centers and partition matrix based on multivariate subsequences data. They also used a reconstruction error as the fitness function of the Particle Swarm Optimization (PSO) algorithm to define a level for detecting anomalies in multivariate data. However, the proposed algorithm cannot reveal the structure of high dimensional multivariate time series due to the issue involved in the PSO algorithm trap in local optima.
Sharma et al.~\cite{sharma2021indoairsense} proposed a low-cost framework named IndoAirSense to estimate and forecast indoor air quality in selected classrooms in the university. They first used Multi-Layer Perceptron (MLP) and eXtream Gradient Boosting Regression (XGBR) to estimate real-time indoor air quality. Then they used LSTM-wF (Long Short Term Memory without using the forget gate) to reduce the complexity of LSTM to forecast indoor air pollutants. Clearly not using the forget gate that keeps the long-term memory, this model could not detect anomalies in the time series dataset.

Mumtaz et al.~{\cite{mumtaz2021internet}} proposed an LSTM-based model for predicting the concentration of different air pollutants to examine the overall quality of an indoor environment. In their research, they collected the base data through IoT sensors that collect different air pollutants (e.g., $NH_3$, $CO$, $NO_2$, $CH_4$, $CO_2$, $PM_2.5$). Their proposed system had the capability of sending alerts after detecting anomalies in the air quality. 

Xu et al.~\cite{xu2019improving} proposed an LSTM model with an added error correction model (ECM) for improving the prediction of indoor temperature in public buildings. In their approach, an ECM is built when the predicted and measured data are co-integrated in the same order, then utilized to revise the prediction of the testing dataset. 
Jung et al.~\cite{jung2021anomaly} utilized LSTM for predicting the conditions of indoor space for facility management based on the three IoT sensor datasets that measure the temperature, humidity, and brightness of a space. LSTM is used to detect anomalous indoor space conditions where the readings on the combination of three indoor air condition datasets deviate from a threshold obtained during training.

Hossain et al.~\cite{hossain2021novel} proposed a combined prediction scheme using two variations of the RNN model (i.e., GRU and LSTM, respectively) to forecast the daily air quality index (AQI) for two of the biggest cities (i.e., Dhaka and Chattogram) in Bangladesh. In their proposal, they used GRU and LSTM as the first and second hidden layers respectively followed by two dense layers as a prediction model. Results reflected that their model followed the actual AQI trends for both cities and demonstrated that their two models improved the overall performance when compared to a single model of GRU or LSTM was used.

Park et al.~\cite{park2018multimodal} introduced a hybrid model of LSTM-VAE (LSTM Variational Autoencoder) to detect multimodal anomalies on robot-assisted feeding systems designed to help people with disabilities often needing physical assistance. They used the hybrid model to analyze the data collected through 155 robot-assisted feeding systems to find any anomalous behaviors exhibited by any robots. Anomalies are found based on the state threshold by calculating a reconstruction score according to the data containing the state of task execution. The result of the ROC curve obtained in their model was higher than other similar approaches. Similarly, Liu et al.~\cite{lin2020anomaly} proposed a combination model VAE-LSTM that combines Variational Autoencoder and LSTM for identifying anomalies that span over multiple time scales. They used VAE to summarise the local information that happens over a short term while using LSTM to analyze the patterns that happen over a longer term which allowed their proposed model to detect anomalies that occur both in short and long periods. 

The time dependency is significantly related to detecting anomalies in the field of Industrial Internet of Things (IIoT) because anomalies happen in both past and current states. In order to detect time series anomalies in IIoT, Wu et al.~\cite{wu2019lstm} proposed an LSTM-Gauss-NBayes approach for detecting anomalies. They first utilized stacked LSTM to deal with time-series data and obtained the prediction error, then used these prediction errors to detect anomalies by the Gaussian Naive Bayes model. The evaluation showed that their proposed method achieved higher accuracy (0.969\%) in Power dataset, compared to the proposals using stacked Bi-LSTM (0.924\%)~\cite{sun2019preheating}, LSTM NN (0.905\%)~\cite{hochreiter1997long}, and MLP (0.873\%)~\cite{pascanu2013difficulty}.

In order to deal with high dimensional anomaly problems in intelligent industrial applications, Homayouni et al.~\cite{zhou2020variational} proposed VLSTM (variational LSTM) to deal with the high dimensional and imbalance industrial big data (IBD). Their VLSTM includes three parts: LSTM encoder, variational reparameterization module, and LSTM decoder. The LSTM encoder and decoder are used to extract the raw input data from high dimensional to low dimensional without losing critical features. The variational reparameterization module was used to deal with reconstructing hidden variables for low-dimensional feature representation by using variational Bayes for network traffic classification. Similarly, Trinh et al.~\cite{trinh2019urban} proposed a hybrid model that combines LSTM with Autoencoder as well as an Isolation Forest. They used LSTM-AE to extract significant features and calculated the reconstruction error. The author used iForest to detect anomalies based on the error vector.

\section{Preliminaries}
\subsection{LSTM}
LSTM stands for long short term memory which is often regarded as an extension of Recurrent Neural Networks (RNN). RNN provided the capability of “short-term memory” which allowed the use of the previous information (at a certain point only) to be used for the present task. Extending from RNN, LSTM architecture provides the capability of “long-term memory” where a list of all of the previous information (opposed to a point of time) is available for the current neural node.

\begin{figure}[h]
	\centering
	\includegraphics[width=1.0\linewidth]{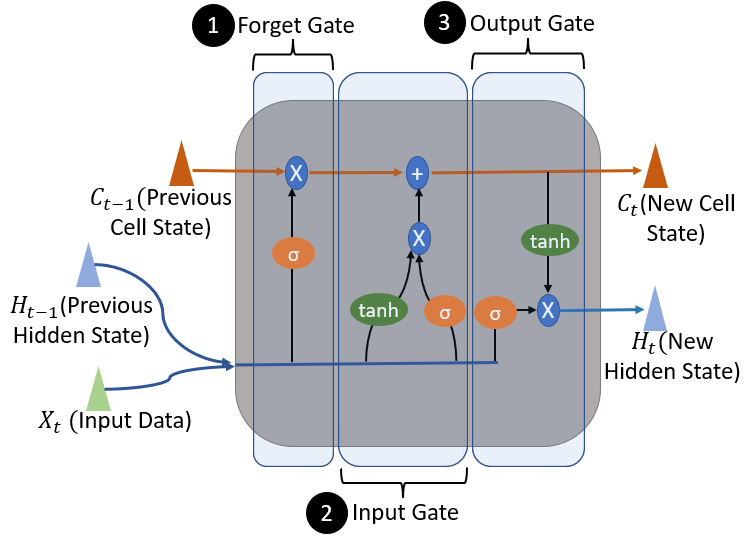}
	\caption{How LSTM Unit Works}
	\label{fig:LSTM_cell}
\end{figure}

A common LSTM unit, depicted in Fig. \ref{fig:LSTM_cell}, is composed of a cell, an input gate, an output gate, and a forget gate. The cell remembers values over arbitrary time intervals and the three gates regulate the flow of information in and out of the cell. 

Note that: 
\begin{itemize}
	\item The \textit{Cell State} refers to the current long-term memory of the network that stores the list of previous information.
	\item The previous \textit{Hidden State} refers to the output at the previous point in time which can be seen as short-term memory.
	\item The input data contains the input value at the current time step.
\end{itemize}

\textbf{Step1 : Forget Gate}\\
The main purpose of the forget gate is to decide which bits of the cell state are useful given both the previous hidden state and new input data. Towards this, the previous hidden state and the new input data are fed into the neural network which generates a vector where each element of the vector is in the interval in the range of [0,1] using a sigmoid activation function. 

The forget gate part of the network is trained so that it outputs close to 0 when a component of the input is irrelevant, otherwise closer to 1 when relevant.  These outputs are then sent up and pointwise multiplied with the previous cell state. Mathematically, the results ($f_t$) from the forget gate can be presented as:
\begin{equation}\label{eq:ft}
	{f_t} = \sigma ({w_f}[H_{t-1},X_t] + {b_f})
\end{equation}
where $\sigma$ is the activation function, $w_f$ and $b_f$ is the wight and bias of the forget gate. $H_{t-1}$ and $X_t$ present the concatenation of hidden state and the current input respectively. 

\textbf{Step2 : Input Gate}\\
The main purpose of the input gate is two folds. The first is to check if the new information (i.e., the previous hidden state and new input data) is worth keeping in the cell state. If there are, secondly, it decides what new information should be added to the cell state. Towards this, the input gate goes through two processes. 

One process involves generating a new memory update vector, represented as $\tilde{C_t}$, by combining the previous hidden state and new input data. A tanh activation function is used to generate the elements of the memory update vector to contain the value in the range of [-1, 1] where the negative values are used to reduce the impact of a component in the cell state. This vector represents how much to update each component of the cell state given the new data. This process is depicted in Equation \ref{eq:tildeCt} as follows.
\begin{equation}\label{eq:tildeCt}
	{\tilde{C_t}} = tanh ({w_c}[H_{t-1},X_t] + {b_c})
\end{equation}
where $w_c$ and $b_c$ are the weight matrices and the bias of the input gate respectively while a tanh is used as an activation function.

Another process of the input gate involves identifying which components of the new input, given the context of the previous hidden state, are worth remembering. Similar to the forget gate, the input gate is trained to output a vector of values in [0,1] using the sigmoid activation function. Any output closer to 0 will not be updated in the cell state. This process is depicted in Equation \ref{eq:it} as follows.
\begin{equation}\label{eq:it}
	{i_t} = \sigma ({w_i}[H_{t-1},X_t] + {b_i})
\end{equation}
where $w_i$ and $b_i$ are the weight matrices and the bias of the input gate.

These two processes are pointwise multiplied. This causes the magnitude of new information decided by Equation \ref{eq:it} to be regulated and set to 0 if needs be. This resulting combined vector is then added to the cell state, resulting in the long-term memory of the network being updated, as shown in Equation \ref{eq:Ct}.
\begin{equation}\label{eq:Ct}
	{C_t} = {{f_t} \odot {C{_t-1}}} + {{i_t} \odot {\tilde{C_t}}}
\end{equation}

\textbf{Step3 : Output Gate}\\
With the updates to the long-term memory done, it is time to work with the output gate. The main purpose of the output gate is to decide the new hidden state. Towards this, the output gates use three different information, the newly updated cell state, the previous hidden state, and the new input data. 

It first applies the previous hidden state and current input data through the sigmoid activated network to obtain the filter vector ${o_t}$ as shown in Equation \ref{eq:ot}.
\begin{equation}\label{eq:ot}
	{o_t} = \sigma ({w_o}[H_{t-1},X_t] + {b_o})
\end{equation}
where $w_o$ and $b_o$ are the weight matrix and the bias of output gate.

The cell state is passed through a tanh activation function to force the values into the interval [-1, 1] to create a squished cell state which is applied to the filter vector by pointwise multiplication. A new hidden state ${H_t}$ is created and outputted, aloing with the new cell state ${C_t}$, as shown in Equation \ref{eq:ht}.
\begin{equation}\label{eq:ht}
	{H_t} = {o_t} \odot tanh({C_t})
\end{equation}

The new cell state ${C_t}$ becomes a previous cell state ${C_{t-1}}$to the next LSTM unit while the new hidden state ${H_t}$ becomes a previous hidden state ${H_{t-1}}$ to the next LSTM unit. These are repeated until the input data from all time series sequences are processed by all LSTM cells involved.

\subsection{Autoencoder (AE)}
\begin{figure}[h]
	\centering
	\includegraphics[width=0.7\linewidth]{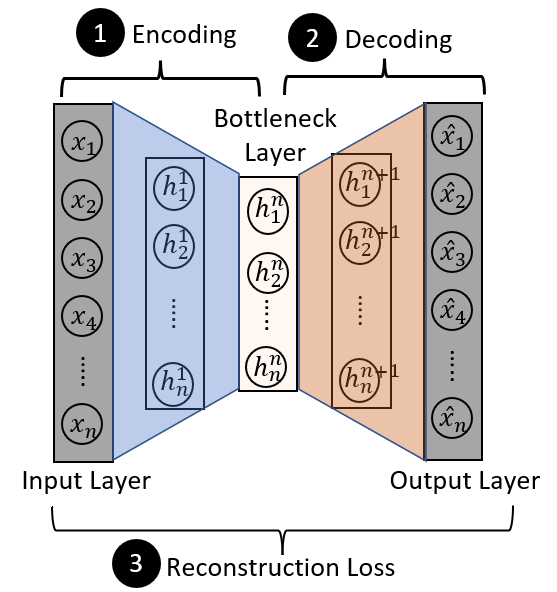}
	\caption{How Autoencoder Works}
	\label{fig:ae}
\end{figure}

An autoencoder is a type of unsupervised neural network that is used to learn efficient codings of unlabelled data. It learns a representation for a set of input data by training the neural network to ignore insignificant data (i.e., often termed as “noise”). A typical autoencoder is composed of an input layer, an output layer, and several hidden layers. The operations of an autoencoder can be divided as Encoding, Decoding, and Reconstruction Loss as illustrated in Fig. \ref{fig:ae}.

\textbf{Step1: Encoding}\\
In the encoding operation, input data $x$ is a $m$ high-dimensional vector $(x \in \Bbb R^m)$ that is mapped to a low-dimensional bottleneck layer representation $(h)$ after removing any insignificant features, as shown in Equation~(\ref{eq:encoder}).
\begin{equation}\label{eq:encoder}
	h= f_1(w_ix + b_i)
\end{equation}
where $w_i$ is the weight matrix, $b_i$ is a bias and $f_1$ is an activation function. \par

\textbf{Step2: Decoding}\\
In the decoding operation, the bottleneck layer representation of $(h)$ is used to generate the output $\hat{x}$ that maps back into reconstruction of ${x}$, as shown in Equation~(\ref{eq:decoder}):
\begin{equation}\label{eq:decoder}
	\hat{x} = f_2(w_jh + b_j)
\end{equation}
where $f_2$ is an activation function for the decoder. $w_j$ is the weight matrix, $b_j$ represents a bias and $\hat{x}$ represent reconstructed input sample. Note that $w_j$ and $b_j$ may be unrelated to the corresponding $w_i$ and $b_i$ for the encoder.

\textbf{Step3: Reconstruction Loss}\\
In a standard autoencoder model, a reconstruction loss $(L)$ is calculated to minimize the difference between the output and the input, as shown in Equation \ref{eq:loss}. It is often this reconstruction loss that is used for anomaly detection task \cite{xu2021improving, wei2021ae}.
\begin{equation}\label{eq:loss}
	\begin{aligned}
		& L(x-\hat{x}) = \frac{1}{n}\sum_{n=1}^{n}|\hat{x}_{t} - x_{t}|\\
	\end{aligned}
\end{equation} 

where $x$ represents the input data, $\hat{x}$ indicates the output data, and $n$ is the number of samples in the training dataset. 

\begin{figure*}[!h]
	\centering
	\includegraphics[width=1.0\linewidth]{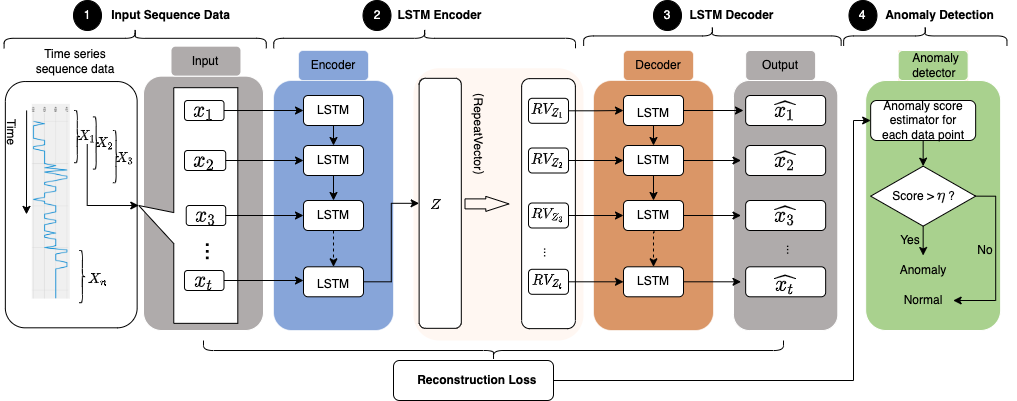}
	\caption{Overview of our proposed model}
	\label{alg:LSTM_AE}
\end{figure*}
However, this is extended to compute a reconstruction loss of a sample in our model as following:
{\begin{equation}\label{eq:loss_time}
		\begin{aligned}
			& x_{i} = \frac{1}{n}\sum_{n=1}^{n}|\hat{x}_{i} - x_{i}|\\
			&\quad \quad s.t. \ \ \  n =
			\begin{cases}
				i & \quad \leq \frac{N+1}{2} \\
				N-i+1 & \quad n > \frac{N+1}{2}
			\end{cases}
		\end{aligned}
	\end{equation} 
}
where $N$ is the total number of samples, and $n$ dedicate the $n$th sample, $X_i=\{x_1,...,x_i\}$.

Then, we compute the  the reconstruction loss for all samples in time-series as follows.
\begin{equation}
	loss =  \frac{1}{N}\sum_{N=i}^1x_i
\end{equation}
where $N$ is the total number of samples, and $x$ indicates a reconstruction loss computed for a sample.

\section{Methodology}\label{sec:method}
In this section, we introduce our proposed model that uses the combination of LSTM and Autoencoder to detect anomalies based on the analysis of time-series data. We first provide the overview of our proposed model based on the four steps of creating the input sequence, LSTM Encoder, LSTM Decoder, and anomaly detection. We also provide a detailed description of the algorithm our model uses in terms of training and testing phases.

\subsection{LSTM-Autoencoder} \label{sec:LSTM_AE}
The overview of our proposed model is illustrated in Fig.~\ref{alg:LSTM_AE}. Our LSTM-Autoencoder utilizes the capabilities of both the LSTM neural network and Autoencoder which builts the LSTM networks on the encoder and decoder schemes of Autoencoder.  The encoder obtains the sequence of the high-dimensional input data as a fixed-size vector. Using the memory cells of LSTM the data processed by the encoder scheme retains the dependencies across multiple data points within a time-series sequence while keeps reducing the high dimensional input vector representation into low-dimensional representation until it reaches the latent space. The decoder LSTM reproduces the fixed-size input sequence from the reduced representation of the input data in the latent space using reconstruction error rates to set a threshold. This threshold is used to detect anomalies.

\textbf{Step1: Input Sequence Data}\\
The original dataset is made as a series of time sequence $[X_1, X_2, X_3,...,X_n]$. Each sequence $X$ with a fixed T-length time window data $[x_1, x_2, x_3,...,x_t]$ is created where $x_t \in R{^m}$ represents an $m$-features input at time-instance $t$. This is then again reshaped into a 2d (2-dimensional) array, representing samples and timesteps. For example, a sequence of our $CO_2$ data is converted into the 2d array where each dimension indicates the list of samples at 10 timesteps.

\begin{figure}[!h]
	\centering
	\includegraphics[width=1.0\linewidth]{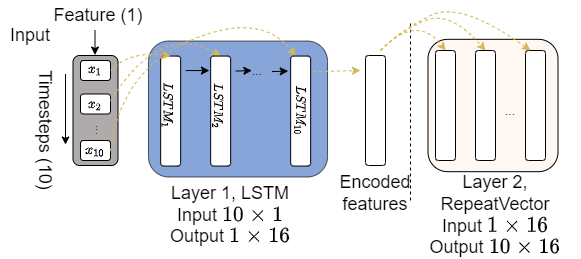}
	\caption{Details of LSTM Encoder}
	\label{fig:lstm_encoder}
\end{figure}

\textbf{Step2: LSTM Encoder}\\
The main purpose of the LSTM encoder is to act like a sequence folding layer that converts features to a batch of time-based feature sequences. It is like convolution operations on timesteps of feature sequences independently. Fig. \ref{fig:lstm_encoder} describes the details of how the (AE) encoder interacts with the series of LSTM unit cells trained to recognize the most relevant features in the input sequence.

Each time series of $X_i$ contains 10 samples collected on 10 timesteps (of a 1-minute interval). This one-dimensional dataset is reshaped into a two-dimensional dataset to feed to the encoder. For example, to unroll the input dataset based on timesteps, the number of input is coveted as a 2d vector where one dimension contains the 10 timesteps and another dimension contains the feature (i.e., samples of $CO_2$ reading), presenting as a vector of $10\times1$. This is now fed to the encoder. 
The encoder creates Layer 1 which contains an LSTM network with 10 LSTM cells. Each LSTM cell unit processes a sample. 10 LSTM cells work in a sequential manner where the 1st LSTM unit passes the result of the sample to the 2nd LSTM. The 2nd LSTM unit decides whether to keep the previous sample from the 1st LSTM to keep or forget it. If the 2nd LSTM decides to keep it, it writes it in the long-term memory and passes the information of the sample from the 1st LSTM along with the feature information processed from the sample it is processing to the 3rd LSTM and so on. The last LSTM, the 10th in our model, has all samples worth keeping processed by the 9 previous LSTM cells. The information about all relevant samples is outputted by the last LSTM cell. This output is now coveted as the $1\times16$ vector as encoded features.

Note that we added a RepeatVector as Layer 2 to create the copies of the $1\times16$ vector as many as equal to the number of timesteps. For example, the size of timesteps in our model is 10 therefore Layer 2 creates 10 copies of encoded features as a two-dimensional vector that equals $10\times16$.

\begin{figure}[!h]
	\centering
	\includegraphics[width=1.0\linewidth]{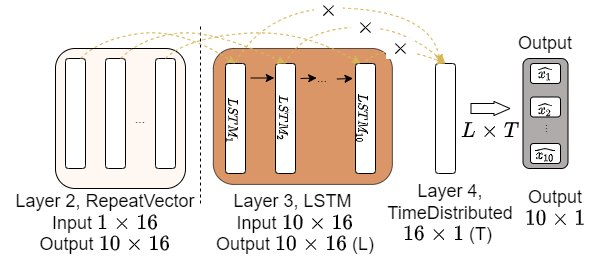}
	\caption{Details of LSTM Decoder}
	\label{fig:lstm_decoder}
\end{figure}

\textbf{Step3: LSTM Decoder}\\
The main purpose of the LSTM decoder is to act like a sequence unfolding layer that restores the sequence structure of the input data after the sequence folding on timesteps. Fig. \ref{fig:lstm_decoder} describes the details of how the decoder interacts with LSTM cells to reconstruct the outputs.

Each $1\times16$ set is now fed as an input to the decoder which creates a Layer 3 network with 10 LSTM cell units. Each LSTM cell unit processes each $1\times16$ encoded feature. Each LSTM unit produces an output that represents the result of the learning from the encoded feature where the output is multiplied with the $1\times16$ vector created by the additional TimeDistribution layer. At the same time, each LSTM cell unit produces another 2nd output containing the state of what has been processed by the current LSTM cell passing to the next LSTM, except the last LSTM unit. Note that matrix multiplication between the output of each LSTM layer ($L$) ($10\times16$) and TimeDistribution layer ($16\times1$) is calculated which results in a vector with the size of $10\times1$ which is the same as the size of the input.

\textbf{Step4: Anomaly Detection}\label{sec:Anormaly}\\
\begin{figure*}[!h]
	\centering
	\includegraphics[width=0.65\linewidth]{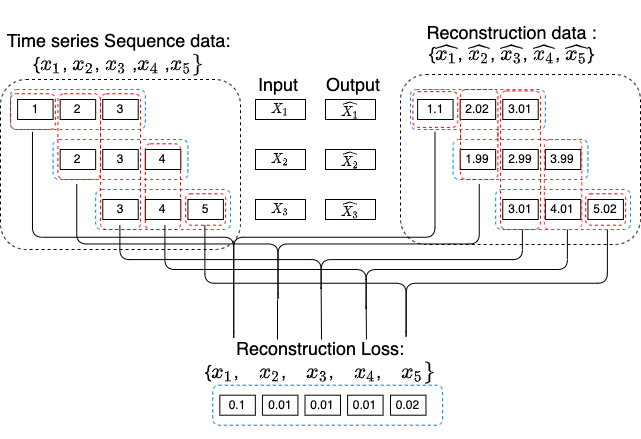}
	\caption{Computing Reconstruction Loss on Time Series} 
	\label{fig:timestamp-mae}
\end{figure*}
An anomaly can be defined as an observation diverging from the majority of the data. A threshold can be set as a decision point to decide how much an observation deviates. Any observations that go beyond the threshold are defined as anomalies. 

Applying this threshold-based anomaly detection technique, our model is trained with the dataset that contains the $CO_2$ values within a normal range. This is to obtain the reconstruction error rates associated with the normal $CO_2$ data points. Once training is done and all different reconstruction error is computed on all samples, the max reconstruction error rate is set as a threshold. Once the threshold is decided, we input the testing $CO_2$ dataset which now contains all ranges of $CO_2$ readings. A reconstruction error rate of each $CO_2$ value is computed for each sample in the testing set. If the reconstruction error rate goes beyond the threshold, this sample is considered as an anomaly.

Fig.~\ref{fig:timestamp-mae} illustrates how we calculate reconstruction loss for each sample contained in different time-series sequences. Let’s presume that there are 5 samples [$x_1, x_2, x_3, x_4,x_5$] which are made as 3 time-series sequences of [$X_1, X_2,X_3$]  where each sequence containing 3 samples on 3 different timesteps where $X_1 \in [x_1, x_2, x_3]$, $X_2 \in [x_2, x_3, x_4$], and $X_3 \in [x_3, x_4, x_5$] - as shown in blue dotted blocks. Our model trains these 3 time-series of sequences as inputs and constructs the outputs that maps to each sequence \begin{math}
	\hat{X_1} \in [\hat{x_1},\, \hat{x_2},\, \hat{x_3}],\, \hat{X_2} \in [\hat{x_2},\, \hat{x_3},\, \hat{x_4}],\, and\: \hat{X_3} \in [\hat{x_3},\, \hat{x_4},\, \hat{x_5}].  
\end{math} 

Let’s assume that the original value for the 3 sequences were:
$X_1 \in [x_1 = 1,\, x_2 = 2,\, x_3 = 3]$, $X_2 \in  [x_2 = 2,\, x_3 = 3,\, x_4 =4]$, and $X_3 \in [x_3 = 3,\, x_4 = 4,\, x_5 = 5]$ where the mapping outputs for each time sequence came out as 

\begin{math}
	\hat{X_1} \in [\hat{x_1} = 1.1,\, \hat{x_2} = 2.02,\, \hat{x_3} = 3.01],\: \hat{X_2} \in [\hat{x_2} = 1.99,\, \hat{x_3} = 2.99,\, \hat{x_4} = 3.99],\: and\: \hat{X_3} \in [\hat{x_3} = 3.01,\, \hat{x_4} = 4.02,\, \hat{x_5} = 5.02].
\end{math} 

The reconstruction loss for each sample can be calculated as:
\begin{align*}
	&x_1 = \lvert 1.1 -1 \lvert \,/\, 1 = 0.1 \\
	&x_2 = (\lvert2.02 - 2\lvert + \lvert1.99 - 2\lvert)\, /\, 2 = 0.01\\
	&x_3 = (\lvert3.01 - 3\lvert + \lvert2.99 - 3\lvert + \lvert3.01 - 3\lvert)\, /\, 3 = 0.01\\
	&x_4 = (\lvert3.99 - 4\lvert + \lvert4.01 - 4\lvert)\, /\, 2 = 0.01\\
	&x_5 = \lvert5.02 - 5\lvert\, /\, 1 = 0.02
\end{align*}
The max reconstruction loss is set as a threshold which in our case is set to be 0.1. During testing, any samples whose reconstruction loss goes beyond 0.1 is now labeled as an anomaly.

\subsection{Algorithm} \label{sec:alg}
The algorithm for the proposed model is shown in Algorithm~\ref{alg:lstm_ae_alg}. The main goal of the training phase in our proposed model is two folds. Firstly, the focus of the training is to minimize the reconstruction error so that the outputs reconstructed from the reduced representation of the input resemble the input as much as possible. Secondly, our model obtains the typical reconstruction error rate associated with the normal range $CO_2$ data points to find an optimal threshold to use for detection during the test phase. The main goal of the testing phase is to use the threshold to detect anomalies in the test dataset.

\begin{algorithm}[!h]
	\SetAlgoLined
	\caption{LSTM-AE Anomaly Detection} 
	\label{alg:lstm_ae_alg}
	\KwIn{\\
		Training set  $\{x_0,x_1,x_2,\dots, x_{n-1}\}$,\\
		Test set $\{x'_0,x'_1,x'_2,\dots, x'_{m-1}\}$,\\
		Timesteps $t$ \\}
	\KwOut{ A Set of anomalies($A_t$) or normal ($N_t$)}
	\Begin{
		\tcc*[f]{\footnotesize Phase 1: To sequence }\\
		$X_i$, $X'_i$: sets of training and testing data based on timesteps (t=10)\\
		\For {$i \in [0, n-t)$}
		{
			$X_i = [x_i::x_{i+t}]$
		}
		\For {$i \in [0, m-t)$}
		{
			$X'_i = [x'_i::x'_{i+t}]$
		}
		\tcc*[f]{\footnotesize Phase 2: LSTM-AE training}\\
		Initialize the parameter of LSTM-AE model (M)\\
		\For{$X_i \in [X_0, X_1, ..., X_{n-t})$ } 
		{
			$\hat{X_i} = M(X_i)$ \\
			$L_{err}$  = $\sum{|X_i -\hat{X_i}|}$ \\
			Update LSTM-AE to minimize $L_{err}$ by Eq.~\ref{eq:loss}
		}
		\tcc*[f]{\footnotesize Phase 3: Threshold setting}\\
		\SetKwFunction{FMain}{RLOSS}
		\SetKwProg{Fn}{Function}{:}{}
		\Fn{\FMain{$X$}}{
			\tcc*[f]{\footnotesize $X_i$ reconstruction error calculation};   
			
			\For{$i \in (0, n-t)$}
			{
				$\hat{X_i} = M(X_i)$ \\
				$Err_{arr}[i, i:i+t] = |\hat{X_i} - {X_i}|$ \\
			}
			\textbf{return} $ Err_{arr}; $ 
		}
		\textbf{End Function}\\
			\tcc*[f]{\footnotesize All data reconstruction error calculation}
			\For{$i \in (0, n)$}
			{
				$l_{arr}[i] = \sum{Err_{arr}[:, i] / \sum(Err_{arr}[:, i] != 0)}$
			}
			return $l_{arr}$
		\\
		\tcc*[f]{\footnotesize Max RLoss from training set}\\
		$ threshold (\eta) = \max(RLOSS([X_0, X_1, ..., X_{n-t}))$
		
		\tcc*[f]{\footnotesize Phase 4: Anomaly detection on testing set}\\
		$ltest_{arr} = RLOSS([X'_0, X'_1, ..., X'_{m-t}))$
		
		\For{$i \in (0, m)$}
		{ 
			\eIf{$ltest_{arr}[i] > \eta$}
			{
				$x'_i \rightarrow A_t$ 
			}
			{
				$x'_i \rightarrow N_t$ 
			}
		}
} 
\end{algorithm}

\textbf{Training Phase}\\
The first step in the training phase is to reshape the original dataset into time-series sequences, as shown in Algorithm~\ref{alg:lstm_ae_alg} \texttt{phase 1: To sequence}. The dataset $X_i$ represents a sequence in the training dataset.  In our model, each sequence contains 10 $CO_2$ samples on 10 timesteps. As depicted in the \texttt{Phase 2: LSTM-AE training} within Algorithm.~\ref{alg:lstm_ae_alg}, the training of the model starts where each sequence is fed to the encoder one at a time where a sample in the sequence is trained by a single LSTM in a sequential manner. Once the training of each sequence completes, the latent space of the encoder rearranges the concatenation of the (relevant) data points as a 1-dimensional encoded feature representation. The RepeatVector layer makes multiple copies of the encoded feature.

The decoder creates an LSTM network with the number of LSTM cells according to the timesteps (i.e., also match the copies of the encoded features). Each encoded feature is processed by a single LSTM cell. The results of the processing by all LSTM cells are made as a single-dimensional vector at the TimeDistributed Dense Layer which produces the output. 
A reconstruction loss between the output and input is calculated, as in the steps 8 to 13. A backpropagation strategy is applied to adjust the weights and parameters of the model. We use Mean Absolute Error (MAE) algorithm, as shown in Equation~\ref{eq:mae}, as the reconstruction error loss function. 
\begin{equation}\label{eq:mae}
	Loss (MAE) = \frac{\sum_{i=1}^{n}\left | x_{i} - \hat{x_i} \right |}{n}
\end{equation}
where $n$ indicates the total number of samples, $x_i$ is the representation of the original input bein feed to the encoder while $\hat{x_i}$ is the output produced by the decoder.

The model trains on all time-series sequences until the reconstruction loss is minimized for all samples. Note that we use 16 neurons in the latent space of the encoder to capture the output from the 10th LSTM. We used "tanh" as the activation function in our proposed model. We also use two Dropout layers (0.2) in the encoder and decoder respectively. We use a RepeatVector layer between the encoder and decoder. An additional TimeDistrutedDense layer is used before the output layer. Once training is complete, the max reconstruction error is obtained as a threshold as shown in  \texttt{Phase 3: Threshold setting.}

\textbf{Testing Phase}\\
The details of the testing phase of our model are shown in the \texttt{Phase 4: Anomaly detection on testing set}. A sequence of time series containing 10 data points of 10 timesteps is fed to the trained LSTM encoder. Note that this time the 10 data points contain all ranges of $CO_2$ values. The LSTM decoder produces a single time series also containing 10 data points of 10 timesteps using the encoded (and reduced) feature representation of the input sample. A reconstruction error rate of each data point is compared with the threshold. The calculation of the reconstruction loss strategy is the same as we mentioned in the training phase in Equation \ref{eq:mae}. If the reconstruction loss value is bigger than the threshold $\eta$, this data point is labeled as an anomaly otherwise labeled as normal. This is shown in the following Equation \ref{eq:anoalies}.

\begin{equation}\label{eq:anoalies}
	X'=
	\left\{ \begin{array}{@{}ll@{}}
		X'_i\ \text{is anomalies}, & \text{if}\ ltest_{arr}[i] > \eta\\
		X'_i\ \text{is normal}, & \text{otherwise}
	\end{array}\right.
\end{equation}
where $X'$ indicates a reconstructed time-series, $X'_i$ is a data point contained in the time-series, and $ltest_{arr}[i]$ is a result from a reconstruction loss function using MAE.

\section{Data and Data processing} \label{sec:re}
We discuss the details of the dataset we use in our study along with the description of the data preprocessing strategies we adopted.

\subsection{Dunedin $CO_2$ Dataset}\label{sec:dataset}
\begin{figure*}[t]
	\centering
	\includegraphics[width=0.9\textwidth]{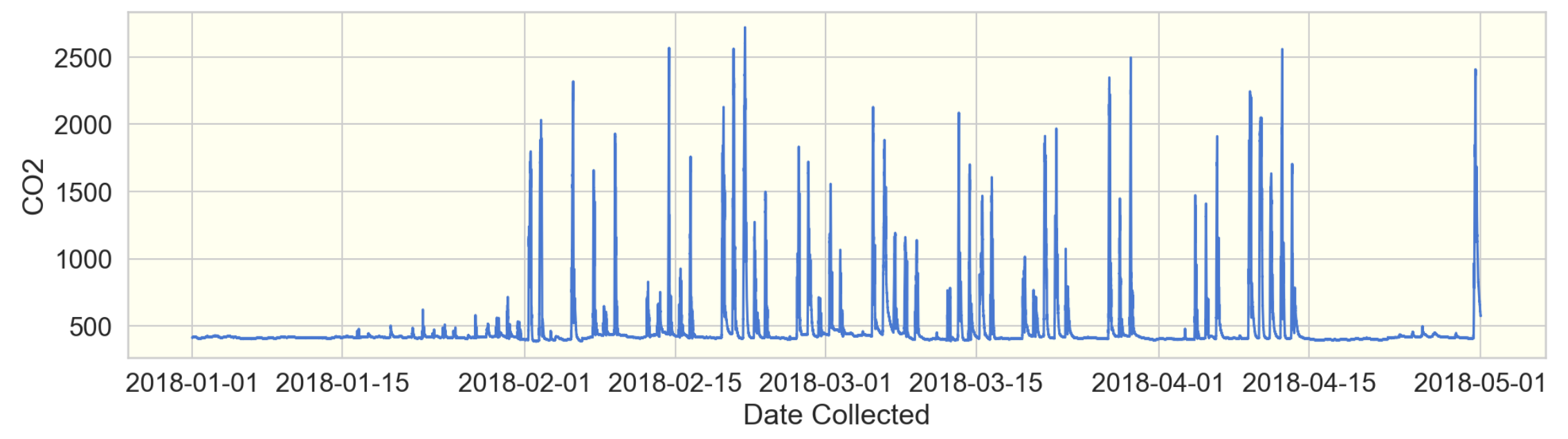}
	\caption{The projection of the original raw dataset}
	\label{fig:org_data_co2}
\end{figure*}

With the focus on understanding the relationship between the level of $CO_2$, weather conditions, and student performance, 74 units of SKOMOBO boxes were deployed in multiple primary/secondary schools in Dunedin, South Island, New Zealand. Records containing the reading of $CO_2$ were collected over a period of four months between 01/01/2018 and 04/30/2018, at a 1-minute interval. The projection of the $CO_2$ reading is shown in Fig. \ref{fig:org_data_co2}. As expected, any changes in the $CO_2$ are not observed when there is a school break (e.g., during January 2018, and the 1st term break in the last two weeks of April 2018. As students occupy classrooms, we observe the changes of $CO_2$ fluctuating a lot of which some could be anomalous. The total number of $CO_2$ readings presented in the dataset was 247,263. 
\subsection{Data Preprocessing} \label{sec:pp}
We first clean up the original records. We first removed all duplicate records. For example, we removed the records containing the $CO_2$ readings with identical timestamps. We also removed the records where it contains both $CO_2$ and timestamp with NaN values. We kept the records where the $CO_2$ reading is either an empty value or NaN value but a timestamp was legit. In this case, we replaced either empty or NaN value with the numeric 0. After this clean-up, we had a total of 171,067 records.

\subsection{Training and Test dataset} 
Two separate datasets for the training and testing phase of our model were prepared as follows.

\textbf{Training Dataset}\\ 
According to \cite{liu2020anomaly}, the typical $CO_2$ value accepted as the normal range are in between 0 and 968 (PPM). 
\begin{figure}[h!]
	\centering
	\includegraphics[width=0.8\linewidth]{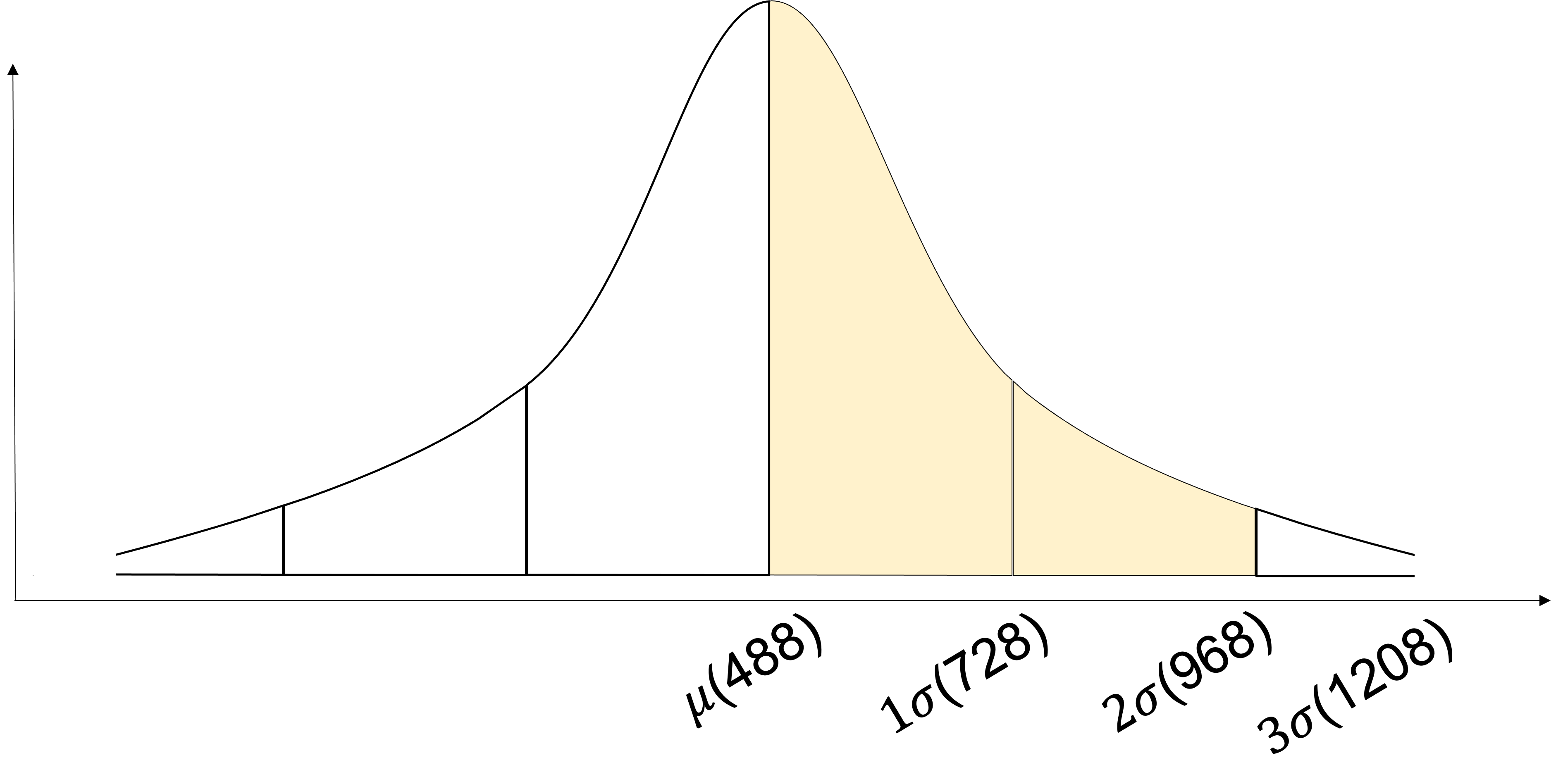}
	\caption{The distribution of $CO_2$ reading according to 3-sigma}
	\label{fig:3-sig}
\end{figure}

Based on our analysis of the dataset, we found that the majority of the $CO_2$ readings sit if we use the 2-sigma rule of the normal distribution (i.e, around $CO_2$ readings less than 968 when the mean of the $CO_2$ readings is around 488) which can be considered as the acceptable normal range, as shown in Fig.~\ref{fig:3-sig}. 

\begin{figure}[h!]
	\centering
	\includegraphics[width=0.98\linewidth]{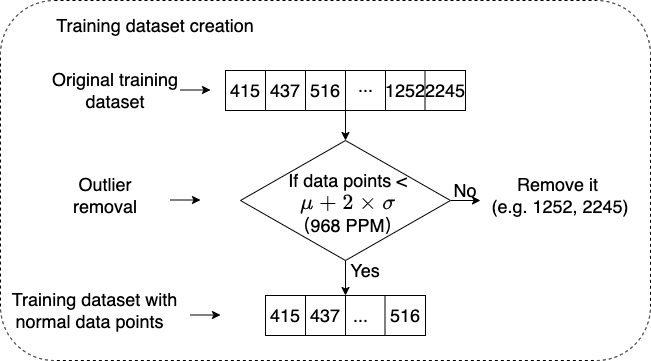}
	\caption{Creating Training Dataset}
	\label{fig:training_data}
\end{figure}

As the training of the model learns the typical reconstruction error associated with normal data points, we created a training dataset only to contain $CO_2$ data points within a normal range. Towards this, we first set aside 3 months of the original dataset (from 01/01/2018 - 31/03/2018). Then we calculated the 2-sigma rule to check each sample if it sits in the normal range or not. If there are any $CO_2$ readings beyond the 2-sigma rule, we removed them. The illustration of creating the training dataset is shown in Fig. \ref{fig:training_data}. 

\begin{figure}[h]
	\centering
	\includegraphics[width=0.98\linewidth]{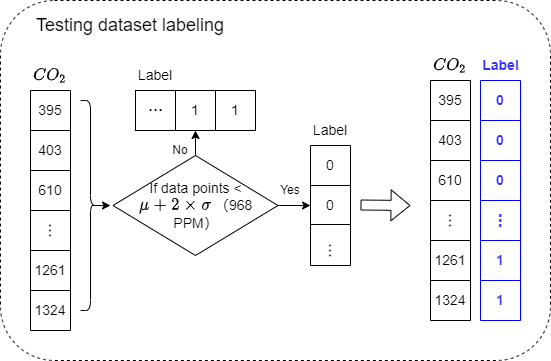}
	\caption{Creating Test Dataset}
	\label{fig:test_data}
\end{figure}

\textbf{Test Dataset}\\ 
We used 1 month of the original data (from 01/04/2018 - 30/04/2018) as a testing dataset. Note that this dataset contains all different ranges of $CO_2$ readings. We added the numeric 0 as a label if the $CO_2$ reading is within the 2 sigma-rule range, otherwise, we added the numeric 1. These labels are only used to evaluate the performance of our model - whether our model was good at detecting anomalous data points to normal data points. The illustration of adding labels in the test dataset is shown in Fig. \ref{fig:test_data}.

\subsection{Data Normalization}
We applied a data normalization technique to eliminate the impacts of different scales across $CO_2$ readings thus reducing the execution time and computational complexity of the model training. We used a standard scalar normalization as depicted in  Equation~(\ref{eq:normalised}).

\begin{equation}\label{eq:normalised}
	Z_i = \frac{X_i-\mu}{S}
\end{equation}

where Z$_i$ denotes all the normalized numeric values ranging between [0-1]; $X_i$ indicates a data point while $\mu$ and $S$ refer to the mean and standard deviation. 
\section{Evaluations}\label{sec:Experiments}
In this section, we provide the details of the experiment including the environment setup,  performance metrics we use, analysis of results, and discussion.

\subsection{Experiment Setup}
Our experiments were carried out using the following system setup shown in Table~\ref{table:Mat}. 

\begin{table}[h]
	\centering
	\footnotesize
	\caption{Implementation environment specification}
	\label{table:Mat}
	\begin{tabular}{p{2.6cm} | p{3.8cm}}
		\hline
		\textbf{Unit}   & \textbf{Description}\\ \hline
		Processor   & 3.4GH$_z$  Inter Core i5 \\ \hline
		RAM  &  16GB      \\ \hline
		OS  &  MacOS Big Sur   11.4  \\ \hline	
		Packages used  &  tensorflow 2.0.0, sklearn 0.24.1    \\ \hline	
	\end{tabular}
\end{table}

The hyperparameters used in the training phase are illustrated with the values for each parameter along with the description in Table~\ref{table:Training parameters}.

\begin{table}[h]
	\centering
	\footnotesize
	\caption{LSTM-AE Training parameters}
	\label{table:Training parameters}
	\begin{tabular}{c|c|c}
		\midrule
		Hyperparameters & Values & Descriptions \\ \midrule
		Learning rate & 0.001 & Learning speed (within range 0.0 and 1.0) \\
		Droupout & 0.2 & No. of neurons ignored \\
		Batch size & 64 & No. of samples in one fwd/bwd pass \\
		Epoch & 30 & No. of one fwd/bwd pass of all samples \\\midrule
	\end{tabular}
\end{table}

\subsection{Performance Metrics}
To evaluate the performance of our model, we used the classification accuracy, precision, recall, and F1 score as performance metrics. Table \ref{table:Matrix} illustrates the confusion matrix.
\begin{table}[h]
	\centering
	\caption{Confusion Matrix}
	\label{table:Matrix}
	\begin{tabular}{| p{2.8cm} | c | p{1.5cm} | p{1.5cm} |}
		\hline
		\multicolumn{2}{|c|}{ \multirow{2}{*}{Total Population} } &   \multicolumn{2}{c|}{Predicted Condition} \\
		\cline{3-4}
		\multicolumn{2}{|c|}{} & Normal & Anomaly \\
		\hline
		\multirow{2}{*}{Actual Condition} & Normal & TN & FP \\
		\cline{2-4}
		&Anomaly & FN & TP \\
		\hline
	\end{tabular}
\end{table}

where;
\begin{itemize}
	\item True Positive (TP) indicates anomalous data point correctly classified as anomalous. 
	\item True Negative (TN) indicates normal data point correctly classified as normal.
	\item False Positive (FP) indicates normal data point incorrectly classified as anomalous.
	\item False Negative (FN) indicates anomalous data point incorrectly classified as normal.
\end{itemize}

Based on the aforementioned terms, the evaluation metrics are calculated as follows: \\
\begin{equation}\label{eq:TPR}
	TPR (True Positive Rate/Recall) = \frac{TP}{TP + FN}
\end{equation}
\begin{equation}\label{eq:FPR}
	FPR (False Positive Rate) = \frac{FP}{FP + TN}
\end{equation}
\begin{equation}\label{eq:PPV}
	Precision = \frac{TP}{TP + FP}
\end{equation}
\begin{equation}\label{eq:F-measure}
	F1-score = 2\times\left(\frac{Precision\times Recall}{Precision + Recall}\right)
\end{equation}
\begin{equation}\label{eq:ACC}
	Accuracy = \frac{TP+TN}{TP + TN + FP + FN}
\end{equation}
\\
The area under the curve (AUC) computes the area under the receiver operating characteristics (ROC) curve which is plotted based on the trade-off between the true positive rate on the y-axis and the false positive rate on the x-axis across different thresholds. Mathematically, AUC is computed as shown in Equation~(\ref{eq:auc}).

\begin{equation}\label{eq:auc}
	AUC_{ROC}=\int_{0}^{1} \frac{TP}{TP+FN}d\frac{FP}{TN+FP}
\end{equation}

\subsection{Results}
We provide the results of the performance of our proposed model observed from a number of different evaluation aspects.
\subsubsection{Training}
\begin{figure}[h]
	\centering
	\includegraphics[width=0.9\linewidth]{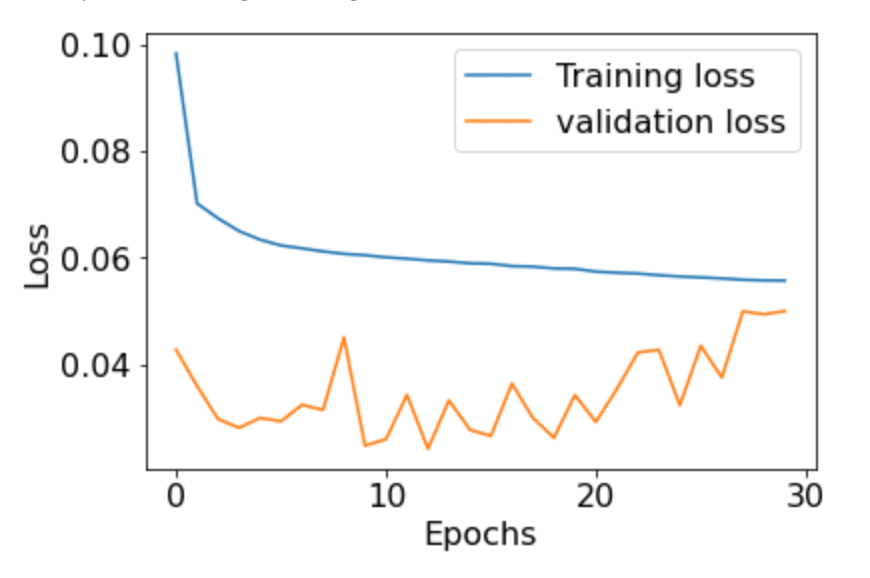}
	\caption{Training/Validation Loss}
	\label{fig:Loss}
\end{figure}

Fig.~\ref{fig:Loss} shows the trends of the loss at different epoch intervals. The training loss (the blue line) assesses the error rate of the model during training. We can see that the training loss stabilizes pretty quickly approximately around 8 epochs. We set aside 10\% of the validation set from the training dataset to assess the performance of our model during training. As expected, the validation loss is not stabilized before 8 epochs. However, after 8 epochs, the validation loss presents a similar loss rate to the training loss (i.e., the average of approximately 0.07\%). This can be regarded as a good fit and our proposed model works well (i.e., our model does not overfit or underfit).

\subsubsection{Impact of Model Architecture}
\begin{figure}[h]
	\centering
	\includegraphics[width=0.9\linewidth]{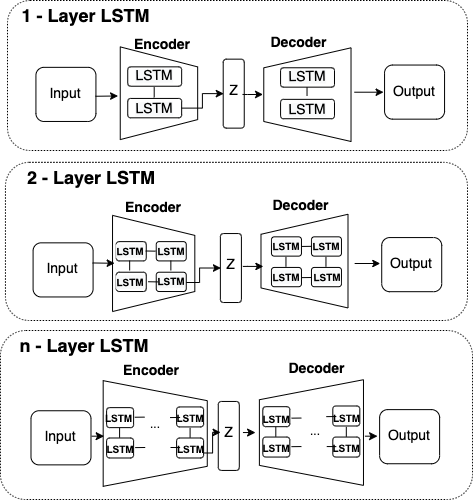}
	\caption{Different Model Architecture}
	\label{fig:model_arch}
\end{figure}

We also tested the sensitivity of our model in terms of the model architecture that differs in the number of the hidden layer (s) and the number of LSTM cell units used. Three different types of model architects, consisting of 1 hidden layer, 2 hidden layers, and $n$ hidden layers, at the encoder and decoder were evaluated. The number of hidden layers at the encoder and the decoder vary while the number of LSTM units used at different model architectures is the same. The details of the three model architectures we evaluated are shown in Fig.~\ref{fig:model_arch}.
There were slight differences in terms of the model performance depending on the number of hidden layers. For example, the model architecture with 1 hidden layer worked best with the F1-score above 94.55\% while the F1-score of 93.48\% and 93.31\% were reached by 2-layer and 3-layer models, respectively. The size of the output vector used by different architectures had a higher impact on the model performance through the smallest size of the output vector used by the 1-layer model worked best reaching 94.55\% F1-score, as shown in Table.~\ref{table:LSTM_AE_compare}.

\begin{table}[h]
	\centering
	\caption{Performance of different model architectures}
	\label{table:LSTM_AE_compare}
	\begin{tabular}{ccccccc} 
		\midrule
		No. (Layers)& No. (Units)& Accuracy & Precision & Recall & F1-score \\ \hline
		1 & 128 & 99.29 & 100 & 85.71 & 92.31\\ \relax
		\textbf{1} & \textbf{16} & \textbf{99.50} & \textbf{100} & \textbf{89.90} & \textbf{94.68}\\ \relax
		2 & 64,16 & 99.39 & 100 & 87.76 & 93.48\\ \relax
		3 & 128,64,16 & 99.38 & 100 & 87.47 & 93.31\\ \midrule
		\end {tabular}
	\end{table}
	
	\subsubsection{Impact of The Size of Time Sliding Window}
\begin{figure}[h]
		\centering
		\begin{tabular}{cc}
			\begin{subfigure}{0.22\textwidth}
				\centering
				\begin{tikzpicture}
					\begin{axis}[
						height=4.5cm,
						width=4.6cm,
						y label style={at={(axis description cs:-0.16,.5)},anchor=south},
						xticklabel style={font=\scriptsize},
						yticklabel style={font=\scriptsize},
						font=\footnotesize,
						legend style={font=\scriptsize ,nodes={scale=0.85},mark options={scale=1}}, 
						xlabel={Time Window Length},
						ylabel={Precision (\%)},
						xmin=0, xmax=8,
						ymin=70, ymax=110,
						xtick={0,1,2,3,4,5,6,7,8},
						ytick={70,80,90,100,110},
						xticklabels={,10,15,20,25,30,35,40},
						legend pos=south east,
						ymajorgrids=true,
						xmajorgrids=true,
						grid style=dashed,
						]
						\addplot[
						color=red!80!black,
						mark=+,
						]
						table [x=Sliding Window Length, y=Precision, col sep=comma]{data2.csv};
					\end{axis}
				\end{tikzpicture}
				\caption{Precision}
				\label{Precision}
			\end{subfigure}
			\begin{subfigure}{0.22\textwidth}
				\centering
				\begin{tikzpicture}
					\begin{axis}[
						height=4.5cm,
						width=4.6cm,
						y label style={at={(axis description cs:-0.16,.5)},anchor=south},
						xticklabel style={font=\scriptsize},
						yticklabel style={font=\scriptsize},
						font=\footnotesize,
						legend style={font=\scriptsize ,nodes={scale=0.85},mark options={scale=1}}, 
						xlabel={Time Window Length},
						ylabel={Accuracy (\%)},
						xmin=0, xmax=8,
						ymin=70, ymax=110,
						xtick={0,1,2,3,4,5,6,7,8},
						ytick={70,80,90,100,110},
						xticklabels={,10,15,20,25,30,35,40},
						legend pos=south east,
						ymajorgrids=true,
						xmajorgrids=true,
						grid style=dashed,
						]
						\addplot[
						black,
						mark=o,
						]
						table [x=Sliding Window Length, y=Accuracy, col sep=comma]{data2.csv};	
					\end{axis} 
				\end{tikzpicture}
				\caption{Accuracy}
				\label{Accuracy}
			\end{subfigure}
		\end{tabular}
		\begin{tabular}{cc}
			\begin{subfigure}{0.22\textwidth}
				\centering
				\begin{tikzpicture}
					\begin{axis}[
						height=4.5cm,
						width=4.6cm,
						y label style={at={(axis description cs:-0.16,.5)},anchor=south},
						xticklabel style={font=\scriptsize},
						yticklabel style={font=\scriptsize},
						font=\footnotesize,
						legend style={font=\scriptsize ,nodes={scale=0.85},mark options={scale=1}}, 
						xlabel={Time Window Length},
						ylabel={Recall (\%)},
						xmin=0, xmax=8,
						ymin=70, ymax=110,
						xtick={0,1,2,3,4,5,6,7,8},
						ytick={70,80,90,100,110},
						xticklabels={,10,15,20,25,30,35,40},
						legend pos=south east,
						ymajorgrids=true,
						xmajorgrids=true,
						grid style=dashed,
						]
						\addplot[
						color=green!80!black,
						mark=triangle,
						]
						table [x=Sliding Window Length, y=Recall, col sep=comma]{data2.csv};	
					\end{axis}
				\end{tikzpicture}
				\caption{Recall}
				\label{Recall}
			\end{subfigure}
			\begin{subfigure}{0.22\textwidth}
				\centering
				\begin{tikzpicture}
					\begin{axis}[
						height=4.5cm,
						width=4.6cm,
						y label style={at={(axis description cs:-0.16,.5)},anchor=south},
						xticklabel style={font=\scriptsize},
						yticklabel style={font=\scriptsize},
						font=\footnotesize,
						legend style={font=\scriptsize ,nodes={scale=0.85},mark options={scale=1}}, 
						xlabel={Time Window Length},
						ylabel={F1-score (\%)},
						xmin=0, xmax=8,
						ymin=70, ymax=110,
						xtick={0,1,2,3,4,5,6,7,8},
						ytick={70,80,90,100,110},
						xticklabels={,10,15,20,25,30,35,40},
						legend pos=south east,
						ymajorgrids=true,
						xmajorgrids=true,
						grid style=dashed,
						]
						\addplot[
						color=blue,
						mark=square,
						]
						table [x=Sliding Window Length, y=F1-score, col sep=comma]{data2.csv};
					\end{axis}
				\end{tikzpicture}
				\caption{F1-score}
				\label{F1-score}
			\end{subfigure}
		\end{tabular}
		\caption{{Performance comparison of LSTM-AE under different time window length}}
		\label{fig:size_time_window}
	\end{figure}
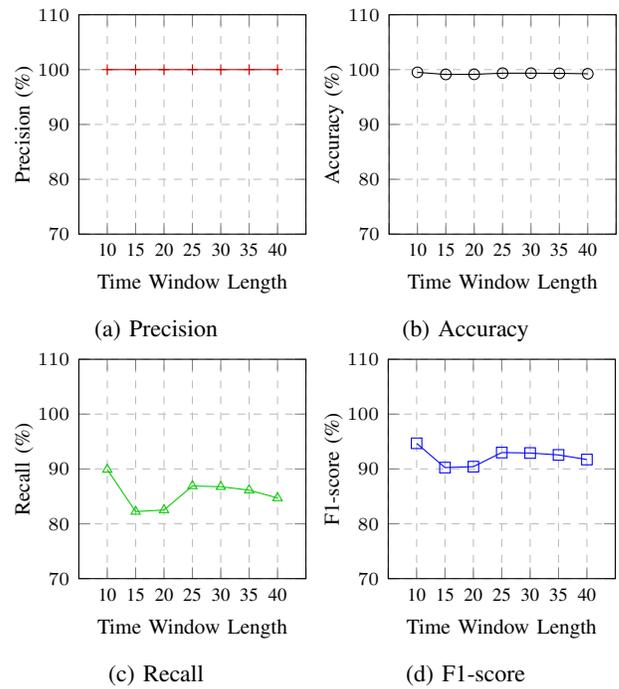

	The size of the time sliding window that decides the number of timesteps to contain in a sequence can impact the overall performance as it can affect the way the reconstruction error rate is computed. Thus we also tested the sensitivity of our model in terms of the size of the time sliding window and the model performance. We tested our model in terms of using the size of the time sliding window at {10, 15, 20, 25, 30, 35, and 40}. As shown in Fig. \ref{fig:size_time_window}, the time sliding window at 10 performed best with the highest TPR rate which contributed to the best F1-score and accuracy. The size of the time window at 15 and 20 worked worst with the lowest TPR rate just above 80\%. Above the time sliding window size 20, we observed that TPR and accordingly f1-score started decreasing as the number of time sliding windows increased. 
	
\subsubsection{Model Performance}
\begin{figure}[h]
	\centering
	\includegraphics[width=0.7\linewidth]{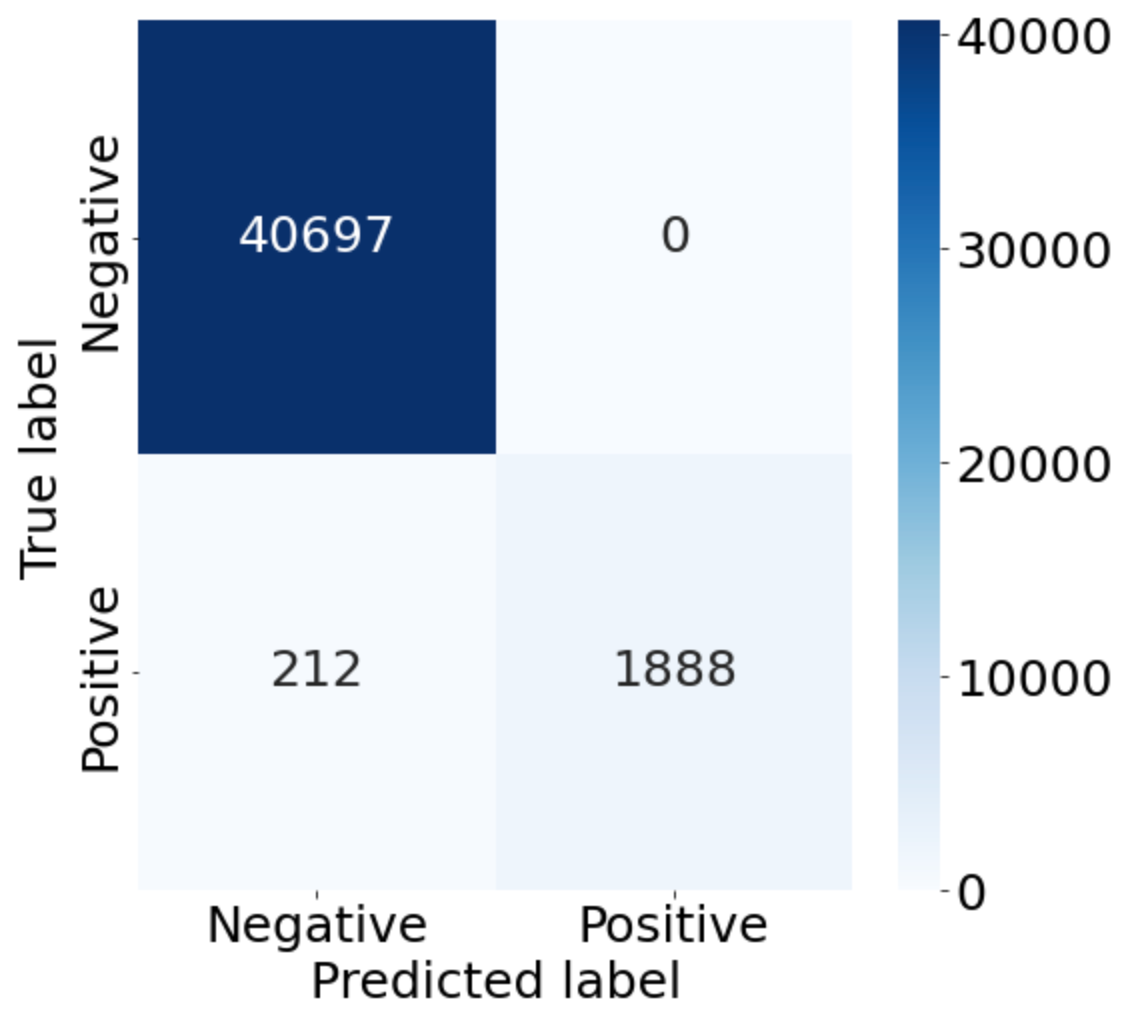}
	\caption{Detection Results Based on Confusion Matrix}
	\label{fig:perf_confusion_matrix}
\end{figure}

Fig.~\ref{fig:perf_confusion_matrix} illustrate the performance of our model based on the confusion matrix. The total number of test samples = 42,787 containing the normal samples = 40,697 and the abnormal samples = 2,100 according to our label. Our model was able to detect a total of 1,888 abnormal data points correctly out of the 2,100 abnormal data samples (i.e., 89.90\% of accuracy). Our model detected the total number of 40,697 normal data points correctly out of the 40,697 normal data samples (i.e., 100\% of accuracy). Our model had none of FP by incorrectly classifying the normal samples as abnormal while it had 212 of FN by incorrectly classifying the abnormal samples as normal. By accounting for all of these, our model resulted in an accuracy of 99.50\%, the precision 100\%, the recall 89.90\%, and the F1-score 94.68\%. 

\begin{figure} [!h]
	\begin{tikzpicture}
		\begin{axis}[
			font=\sf,
			xlabel={False Positive Rate},
			ylabel={True Positive Rate},
			xmin= -0.05, xmax=1,
			ymin= 0, ymax=1,
			xtick={0,.2,.4,.6,.8,1},
			ytick={0,.2,.4,.6,.8,1},
			legend pos=south east,
			legend image post style={scale=0.2},
			no markers,
			]
			\addlegendimage{empty legend}
			\addlegendentry{\textbf{AUC}}
			
			\addplot[smooth,red, tension=0.1] 
			coordinates {(0,0)(0.00,0.8990)(1,1)}; 
			\addlegendentry{$T_{10}$ =  95.0}
			
			\addplot[smooth,green, tension=0.1] 
			coordinates {(0,0)(0.00,0.8224)(1,1)}; 
			\addlegendentry{$T_{15}$ =  91.1}
			
			\addplot[smooth,purple, tension=0.1] 
			coordinates {(0,0)(0.00,0.8252)(1,1)}; 
			\addlegendentry{$T_{20}$ =  91.3}
			
			\addplot[smooth,blue, tension=0.1] 
			coordinates {(0,0)(0.00,0.8690)(1,1)}; 
			\addlegendentry{$T_{25}$ =  93.5}
			
			\addplot[smooth,yellow, tension=0.1] 
			coordinates {(0,0)(0.00,0.8676)(1,1)}; 
			\addlegendentry{$T_{30}$ =  93.4}
			
			\addplot[smooth,pink, tension=0.1] 
			coordinates {(0,0)(0.00,0.8614)(1,1)}; 
			\addlegendentry{$T_{35}$ =  93.1}
			
			\addplot[smooth,pink, tension=0.1] 
			coordinates {(0,0)(0.00,0.8471)(1,1)}; 
			\addlegendentry{$T_{40}$ =  92.4}
			\addplot[black,dashed] coordinates{(0,0) (1,1)};
		\end{axis}
	\end{tikzpicture}
	\caption{AUC-ROC Visualization}
	\label{fig:AUC_ROC}
\end{figure}
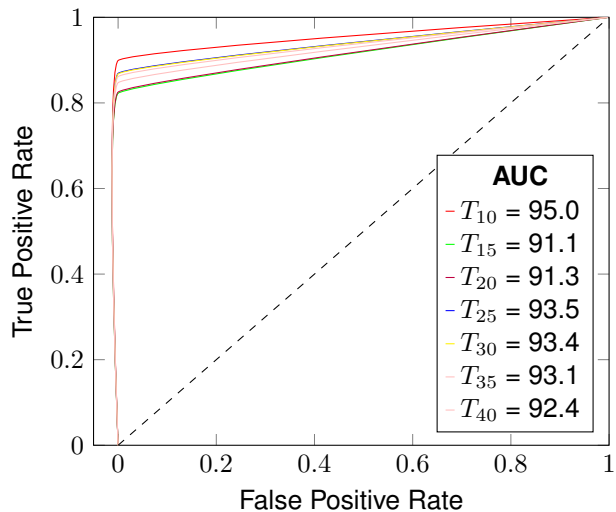

Fig.~\ref{fig:AUC_ROC} shows our model performance in terms of the AUC-ROC graph that clearly demonstrates the trade-off between true positive rate and false-positive rate. The curve confirms that our proposed model is highly effective in accurately detecting anomalies by achieving an AUC-ROC score of 94.8\%. This result is calculated based on the whole time series testing dataset. To detect the model efficiency, we also compared different time window lengths in 1 minutes intervals. We observed that the AUC-ROC curve starts decreasing as the size of the time window length increases. The best performance was shown at 94.8\% when the size of the time window length = 10 while the worst performance was shown when the size of the time window length was = 10 and 20. Similar to the performance of the impact of the time sliding window, the AUC-ROC score decreased slightly as the size of the time window increased from the size over 25.

\begin{figure*}[h!]
	\centering
	\includegraphics[width=0.7\linewidth]{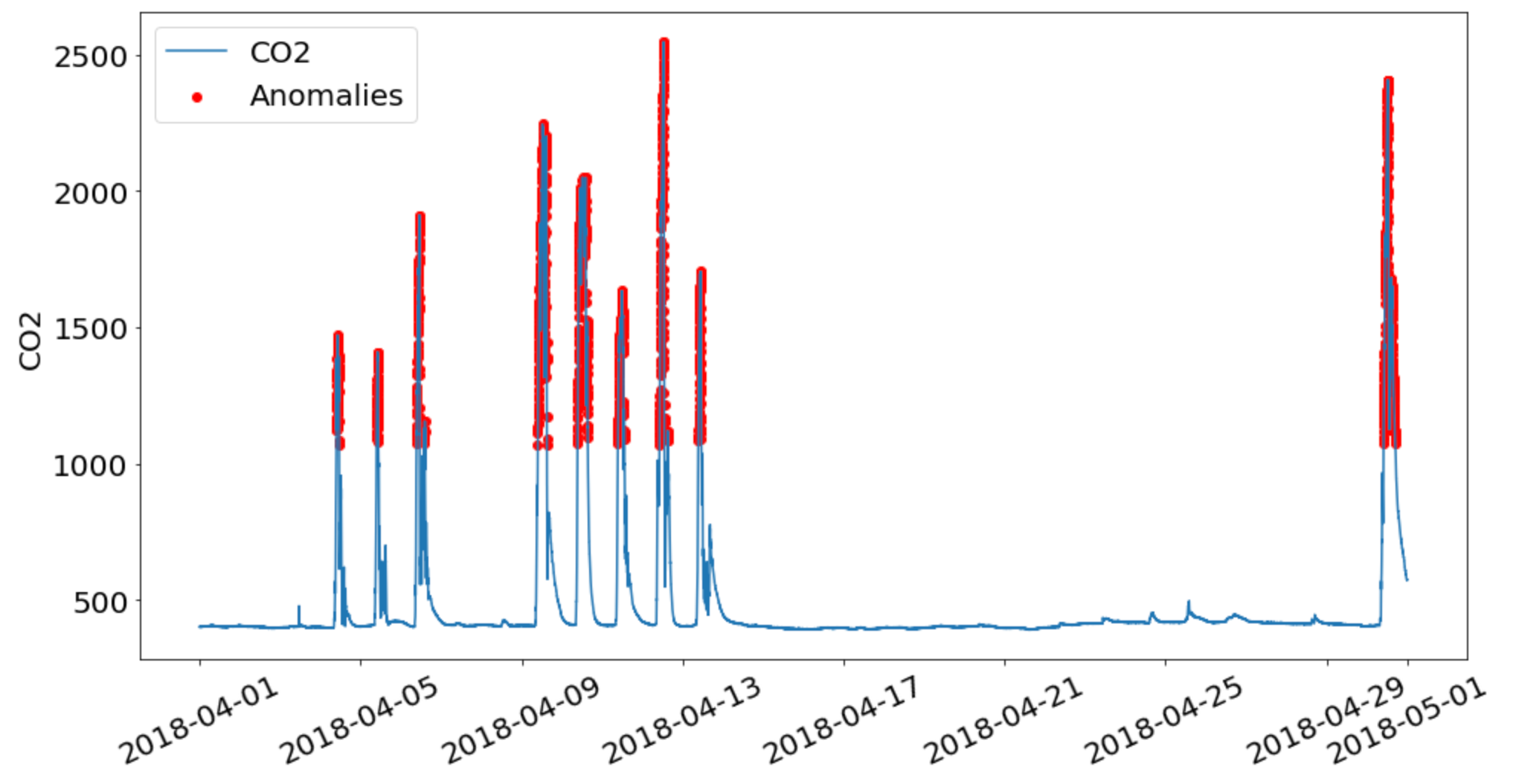}
	\caption{Normal and anomalies distribution on the testing set}
	\label{alg:anomaly_1}
\end{figure*}

Fig.~\ref{alg:anomaly_1} shows where anomalous data points were detected (i.e., see by the red dots) based on the test dataset. In our observation, the threshold was set at 1.742 when the training was done. During the test, we also observed that any $CO_2$ readings greater than 1,000 usually had a reconstruction error rate greater than 1.742 and therefore were considered anomalous. 

\subsubsection{Comparison to Other Similar Models}
Table~\ref{table:Comparison} shows the performance comparison of our model evaluated on the Dunedin $CO_2$ dataset with other similar models that use different variations of LSTM Autoencoder. As the result shows, our approach shows the best performance in terms of both accuracy (at 99.50\%) and precision (at 100\%). The similar model proposed by Yin et al.~\cite{yin2020anomaly} shows the most competitive performance compared to ours with similar accuracy and F1-score. The model proposed by Nguyen et al.~\cite{nguyen2021forecasting} shows a higher F1-score (at 96.98\%) though the accuracy of their model is lower than our method. In our further investigation, they used a One-Class SVM as an additional classifier to reduce the false positives. 

\begin{table*}[!h]
	\centering
	\caption{Comparison to Other Similar Models}
	\label{table:Comparison}
	\begin{tabular}{ccccccc}
		\midrule
		Paper      & Techniques     & Datasets  & Accuracy & Precision & Recall  & F1-score \\ \hline
		Yin et al.~\cite{yin2020anomaly} & LSTM-AE & Yahoo Webscope S5 & 99.25 &   97.84  & 94.16 & 95.97 \\ 
		Liu et al.~\cite{liu2022arrhythmia}  & LSTM-AE &  ECG   & 98.57 & 97.55  &    97.55  & -	\\ 
		Chander et al.~ \cite{chander2021auto} & LSTM-AE &  WSNs position at IBRL    & - &   89.1    &    86.9  & 85.24	\\ 
		\multirow{4}{*}{Lin et al.~\cite{lin2020anomaly}} & \multirow{4}{*}{VAE-LSTM} & Ambient temperature & - & 80.6 & 1.0 & 89.2 \\
		&  & CPU utilization AWS & - & 69.4 &1.0  & 81.9 \\
		&  & Machine teperature & - & 55.9 & 1.0 & 71.7  \\
		Sharma et al.~\cite{sharma2020user} & LSTM-AE &  CERT insider threat dataset  & 90.17&  -  & 91.03& - \\ 
		\multirow{2}{*}{Kieu et al.~\cite{kieu2018outlier}} & \multirow{2}{*}{LSTM-AE} & Numenta Anomaly Benchmark (NBA) & - & 90.8 &98.8  & 94.6 \\
		\multirow{2}{*}{Li et al.~\cite{li2019vidanomaly}} & \multirow{2}{*}{LSTM-AE adversarial learning} & e-VDS &94.16 & 90.31 &88.45  & 89.37 \\
		& & CCV & 83.03 &75.08 & 73.26 & 74.16 \\ 
		Kang et al.~\cite{kang2021anomaly}& LSTM-AE&Break Operating Unit (BOU) data & 94.44 & 97.94 & 85.77& 91.45 \\
		Nguyen et al.~\cite{nguyen2021forecasting} & LSTM-AE-OCSVM & Generated dataset  & 98.36 & 98.45  & 99.59 & 96.98	\\ 
		Tran et al.~\cite{tran2020anomaly} & LSTM-AE iforest &  simulated data in Fashion industry & 95 & 100 & 94 & 87 \\ 
		Our Proposal & LSTM-AE & Dunedin $CO^2$ Dataset & \textbf{99.50} & \textbf{100} & \textbf{89.90}& \textbf{94.68}   \\ \midrule
	\end{tabular}
\end{table*}

\section{Conclusion}\label{sec:Conclusion}
We proposed an LSTM-Autoencoder based deep-learning technique for detecting anomalies in indoor air quality datasets. In our proposed model, two LSTM networks each of which consists of multiple LSTM units provide the learning ability to identify long-term correlational dependencies that exist in a time series sequence. Autoencoder is used to generate encoded features of the input representation while maintaining the long-term dependences identified by the LSTM encoder and constructing the outputs to resemble the input through the LSTM decoder. The max MAE from the trained model on the training set is set as a threshold and is used by the anomaly detector. The anomaly detector identifies each data observation from the testing set as anomalies where its reconstruction loss result is greater than the threshold. 

Our proposed model was applied to the CO$_2$ time-series dataset obtained from a real-world deployment. The experimental results showed that our proposed model is highly efficient for anomalous CO$_2$ readings by providing the detection accuracy of 99.50\% and outperforms other similar models.

We plan to apply our proposed model for detecting DDoS attacks \cite{wei2021ae} based on the time-series analysis.

\bibliographystyle{IEEEtran}
\bibliography{mybib}

\end{document}